\documentclass[runningheads]{llncs}
\usepackage{booktabs}
\usepackage[numbers,sort]{natbib}
\makeatletter \renewcommand\@biblabel[1]{#1.} \makeatother % change format from [1] to 1.
\usepackage[T1]{fontenc}
\usepackage{graphicx}
\usepackage{algorithm}
\usepackage{algorithmic}
\usepackage{caption}
\usepackage{adjustbox}
\usepackage{csvsimple}
\usepackage{pgfplots}
\usepackage{pgfplotstable}
\pgfplotsset{compat=1.17}
\usepackage[absolute,overlay]{textpos}
\usepackage{multirow}
\usepackage{makecell}
\usepackage{subfig}
\usepackage{comment}

\begin{document}
\title{Solving 7x7 Killall-Go with Seki Database}
% Solving 7x7 Killall-Go with Seki Database
\author{Yun-Jui Tsai\inst{1}\orcidID{0009-0007-6703-1687} \and
Ting Han Wei\inst{2}\orcidID{0009-0004-6060-1905}
Chi-Huang Lin\inst{1}\orcidID{0009-0000-5078-7866} \and
Chung-Chin Shih\inst{3}\orcidID{0000-0003-4261-4871} \and 
Hung Guei\inst{3}\orcidID{0000-0002-5590-7529} \and
I-Chen Wu\inst{1}\orcidID{0000-0003-2535-0587} \and \\
Ti-Rong Wu\inst{3}\orcidID{0000-0002-7532-3176}
}
\authorrunning{Tsai et al.}
\institute{National Yang Ming Chiao Tung University, Hsinchu, Taiwan \and
Kochi University of Technology, Kami City, Japan
% \email{lncs@springer.com}
\\
% \url{http://www.springer.com/gp/computer-science/lncs} 
\and
Academia Sinica, Taipei, Taiwan
\\
\email{tirongwu@iis.sinica.edu.tw}
}

\maketitle              % typeset the header of the contribution
\begin{abstract}
Game solving is the process of finding the theoretical outcome for a game, assuming that all player choices are optimal. 
This paper focuses on a technique that can reduce the heuristic search space significantly for 7x7 Killall-Go.
In Go and Killall-Go, \emph{live} patterns are stones that are protected from opponent capture. 
Mutual life, also referred to as seki, is when both players' stones achieve life by sharing liberties with their opponent. Whichever player attempts to capture the opponent first will leave their own stones vulnerable. 
Therefore, it is critical to recognize seki patterns to avoid putting oneself in jeopardy.
Recognizing seki can reduce the search depth significantly.
In this paper, we enumerate all seki patterns up to a predetermined area size, then store these patterns into a seki table. This allows us to recognize seki during search, which significantly improves solving efficiency for the game of Killall-Go.
Experiments show that a position that could not be solved within a day can be solved in 482 seconds with the addition of a seki table.
For general positions, a 10\% to 20\% improvement in wall clock time and node count is observed.

\begin{comment}
Game solving is a challenging task pursued to ensure the correctness of players' every decision. 
In prior work, Proof Cost Network (PCN) was proposed to replace AlplaZero as the heuristic to support proof number search and make the solving process much more efficient. 
However, in Go or Killall-Go, the coexistence area is a problematic part of solving. Seki area is a coexistence area in which neither player can capture the opponent’s stone. The winning condition for White in KIllall-Go is to reach Benson or Seki, but if the Seki condition is not determined, once the solver gets into a situation with Seki, the game length will be extraordinarily extended and consequently increase the depth and complexity of solving the procedure.  
In this paper, we propose a Seki table that is generated considering all the permutations within a given area size. 
With the Seki table, the solver will immediately return the White win once the board matches the Seki table. 
In the experiment, the base solver will be stuck by a problem with Seki situation and remain unsolved even after a day. While with the Seki table, the same problem can be solved in thousands of seconds.
\end{comment}

\end{abstract}

\section{Introduction}
Games solving \cite{vandenherik_games_2002}, particularly for the complex game of Go, is one of the most challenging pursuits in artificial intelligence.
While AlphaZero \cite{silver_general_2018} has mastered 19x19 Go in game playing, only up to 5x6 Go has been fully solved \cite{vanderwerf_solving_2009}.
One interesting variant of Go is \textit{Killall-Go}, which follows similar rules but with Black aiming to capture all White's stones to win.
The game can also be viewed as a whole board \textit{life-and-death} problem in Go, which is a fundamental concept for Go learners.
Killall-Go is therefore a valuable test bed for solving larger board sizes in Go, with many attempts to solve the 7x7 version of the game \cite{shih_novel_2022, wu_alphazerobased_2021, wu_game_2024}.

\begin{figure}[h]
    \captionsetup[subfigure]{justification=centering}
    \centering
    \subfloat[A Killall-Go opening.]{
        \includegraphics[width=0.2\columnwidth]{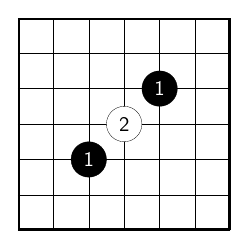}
        \label{fig:killallgo-example}
    }
    \subfloat[White wins by Benson safety.]{
        \includegraphics[width=0.2\columnwidth]{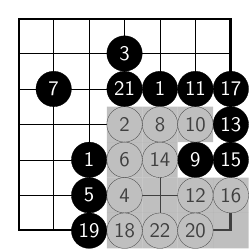}
        \label{fig:benson-example}
    }
    \subfloat[White wins by seki.]{
        \includegraphics[width=0.2\columnwidth]{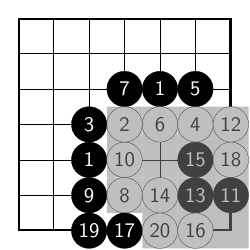}
        \label{fig:seki-example}
    }
    \caption{An illustration of a 7x7 Killall-Go opening with two different winning conditions for White.}
    \label{fig:killallgo}
\end{figure}

In 7x7 Killall-Go, Black plays two consecutive moves first, followed by alternating turns between White and Black, as shown in Fig. \ref{fig:killallgo-example}.
There are two ways for White to win: by achieving unconditional life (identifiable via the Benson algorithm \cite{benson_life_1976}), shown in Fig. \ref{fig:benson-example}, or by reaching mutual life with Black, known as \textit{seki}, shown in Fig. \ref{fig:seki-example}.
In comparison to knowledge-based analysis for Benson safety, seki requires an exhaustive search to identify.
Knowing when to attempt detecting seki to minimize overhead costs is an issue that has yet to be addressed. 
Additionally, by pre-computing seki information, we can save significant time during the game solving process.
For this reason, we propose constructing a seki database, using the detection method proposed by Niu et al. \cite{niu_recognizing_2006}, to aid with solving 7x7 Killall-Go.
In the best case, when using this seki database, positions that cannot be solved in a day can be solved in just 482 seconds.
Our experiments also show that the addition of the database also improves the overall solving time for general cases, with a 10\%-20\% reduction in search time.

\section{Background}

\subsection{Game Solver}
\label{subsec:background_solver}
\begin{comment}
(game solving, PNS, PCN, Application of PCN)
- introduce what is game solver?
    - ting: checkers proof?
    - ting: van der werf's 5x6?
- introduce current game solvers:
\end{comment}

A game is considered solved when its game-theoretic value is found, i.e. we know the outcome under optimal play.
%For two player games with a winning and a losing side, an optimal strategy for both players, known as the solution tree, contains at least one winning move for all winner turns, and all possible moves for all loser's turns.
Since the search space is often extremely large, heuristics are often used to guide the search, minimizing the number of winning moves explored, while simultaneously searching through the shortest game length that leads to a solution.
AlphaZero-like algorithms are known for producing strong agents that do not necessarily attempt to finish games as quickly as possible \cite{wu_alphazerobased_2021}, which make them less ideal for game solving.
Proof-number search (PNS) \cite{allis_proofnumber_1994}, depth-first proof number search (DFPN) \cite{kishimoto_dfpn_2003}, and threat-space search \cite{allis_gomoku_1994} are some common search algorithms that prune unnecessary branches when solving games, potentially leading to more efficient solutions.  

A notable example of a game solved is checkers.
Schaeffer et al. \cite{schaeffer_checkers_2007} used a distributed solving system comprised of a proof-tree manager and numerous workers.
The manager breaks the problem down into tasks, consisting of interesting game positions, which are sent to workers.
The workers were each an instance of a solver with a strong checkers playing program providing heuristic value.
Once a worker finds a solution for a game position, it returns the result to the manager, which uses that information to construct a solution tree.
In addition to this process -- referred to as the forward search -- they also computed a large collection of endgame databases.

We use the online fine-tuning distributed solver \cite{wu_game_2024} in this paper. 
This system follows the manager-worker paradigm.
Each solver is a Monte-Carlo tree search (MCTS) solver \cite{winands_montecarlo_2008} that uses a Proof Cost Network (PCN) \cite{wu_alphazerobased_2021} to provide heuristics, the Benson \cite{benson_life_1976} algorithm to determine terminal conditions, and the Relevance-Zone based search \cite{shih_novel_2022, shih_localpattern_2023} to prune irrelevant nodes. 
% Rather than relying on a strong playing program to provide the heuristic value, this system trains a Proof Cost Network (PCN) \cite{wu_alphazerobased_2021} to estimate the effort needed to solve a game position and guide the solving process.
% In addition, to better address the out-of-distribution positions that do not appear during training, this work applies a fine-tuning method.
During the solving process, the online trainer continuously fine-tunes the deep learning-based heuristic to maintain its accuracy. 
Additionally, the seki database proposed in this paper can also be thought of as endgame information that can reduce the search space significantly.

\subsection{7x7 Killall-Go}
%(killallgo rule, killallgo wining condition, benson)

Killall-Go is a two-player, zero-sum game like Go. In 7x7 Killall-Go, Black is given a large advantage by placing two stones in their first turn. Accordingly, Black is expected to capture all White stones to win. On the other hand, White only needs to secure one safe area to win. 
To determine whether an area is secure, Benson \cite{benson_life_1976} proposed an algorithm based on Go rules to determine whether a set of blocks is unconditionally alive (UCA), i.e. the block is guaranteed to be safe from capture, even if the opponent is allowed an unlimited number of consecutive turns. 
There are two core rules in Go. First, a string of connected stones are called \textit{blocks}, and empty grids that are adjacent to blocks are called \textit{liberties}. A block is captured, with its stones removed from the board, when it no longer has any liberties.
Second, neither player is allowed to capture stones of their own.
With these two rules in mind, a block is UCA if it has at least two liberties in which their opponent may not play in.
Benson's algorithm examines blocks systematically to determine if this is true.
It is worth noting that UCA is a strong guarantee.
White does not need to achieve UCA to secure a safe area to win in Killall-Go.
In fact, Black and White can coexist in the same area, sharing liberties between their stones, unable to capture each other. This situation is referred to as mutual life, or \textit{seki}. 

\subsection{Seki}
\label{subsec:seki}

In Killall-Go, seki often involves 1) White securing an area; 2) Black occupying the boundary of the White area, while also attempting to capture white stones inside it, as shown in Fig. \ref{fig:seki-example}.
In the seki area, neither Black nor White can capture all opponent stones, nor achieve UCA.
In fact, whichever player plays inside the seki area renders their stones vulnerable for capture. 
Thus, players can only move outside the seki area or pass when playing optimally. 

A seki situation signifies secure territory, which in turn means White has won.
However, if the search cannot recognize seki, White must satisfy the stronger condition of UCA to win. Therefore, the only way to arrive at this conclusion is for Black to play inside the seki area, which might not occur until much deeper in the search because it is a suboptimal move.

In addition to the local seki described above, there are also global seki, where the shared liberty is not enclosed. We focus on local seki in this paper, because global seki are difficult to detect and are much rarer in 7x7 Killall-Go.

%Seki can be categorized into local and global instances. Local seki is recognized under the condition that the area outside the seki is assumed to be completely occupied by the opponent's stones, with no liberties outside. The outcome of a local seki may change if the boundary liberty situation is altered. Meanwhile, global seki considers all stones on the board.

%(Seki related research for recognizing)

Previously, Niu et al. \cite{niu_recognizing_2006} describe the issue of recognizing seki thoroughly and propose algorithms for recognizing global and local seki.
Niu's local seki algorithm takes a region as input and generates all legal moves in the region, including passes. The region is searched twice using DFPN, where Black or White play first. Where Black plays first, if the result is a win for Black, the situation is not a seki. Otherwise, if the result is a loss for Black, the situation can either be seki or a White win. The second search assumes White goes first. If the result is a White win, the situation is determined to not be seki. Otherwise, the situation is confirmed to be seki.
We omit the more complicated global seki detection method in this paper.
% The position is first split into several areas, where the local seki detection algorithm is applied to each area. 
% After each local seki search, new information is updated as external information for adjacent areas.
% If, after examining every area, no new information is updated, the global seki detection algorithm ends.

Gol’berg et al. \cite{gol’berg_combinatorial_2014} propose projecting the positional information onto a matrix, which is composed of the shared liberties in the seki. They then describe mathematical conditions that need to be satisfied to confirm a seki. 
Wolf \cite{wolf_seki_2017} proposes a graph representation that forms a topological description of seki. However, its usage is limited to situations where all blocks have two liberties.
Wolf \cite{wolf_two_2007} also introduces a computer program called GoTools, which includes life or death analysis, and a large database of single eye patterns.
However, to our knowledge, the database and methods from GoTools have not been extended for seki detection nor game solving. 

% TODO?
Kishimoto and M{\"u}ller \cite{kishimoto_search_2005} built the program TSUMEGO EXPLORER to solve life and death problems like GoTools, which can also be used to analyze seki. While they emphasize general methods such as DFPN, they also propose heuristics such as the miai strategy and forced moves.
In addition, M{\"u}ller \cite{martin_playing_1997} designed a set of static rules that can be used with a search to recognize safe areas in Go earlier than the Benson algorithm.
We do not use these heuristics and static rules in this paper.

Lastly, several efforts were made to classify safe patterns instead of search. 
Vil{\'a} \cite{vilà_when_2004} proposes identifying single eye shapes to help with game solving, discussing how different eye shapes affect safety in detail.
Cazenave \cite{cazenave_generation_} generates a pattern database for Go, focusing on the pattern's external condition. 
% More specifically, a wide variety of patterns with different external liberties are generated and saved.
Adding external conditions enables each pattern to capture a wider range of board states without increasing the complexity of the search tree.
The motivation is similar to this paper, but the database in this paper does not consider the external conditions of patterns.

% There have been several works \cite{martin_playing_1997, vilà_when_2004, cazenave_generation_} for recognizing safe territories in Go solving.
% % TODO
% The recognition of seki area is also related to recognizing secure territories \cite{martin_playing_1997, vilà_when_2004, kishimoto_search_2005}. 
% Mueller tries to recognize the secure areas in go to reach the terminal condition earlier than the Benson. 
% Kishimoto et al. build apply a high-performance search engine in the Life and Death problem of go. 

\section{Method}
\begin{comment}
- an algorithm for finding seki pattern in Killall-Go

- Method
    - method of generating all permutation
        -algorithm
        -limitation: only can generate pattern with one area
    - method of recognizing seki pattern
        -algorithm
        -only for local seki
    - storing into the table 
    - usage of the table
        - table in the game solver
        - query circumstance
\end{comment}

In this section, we describe how the seki database is created and how it is used.
First, we enumerate all potential seki patterns for specific area sizes. 
Second, each pattern is analyzed via exhaustive search to determine whether they are seki, where valid entries are stored in the database.
Lastly, we describe how the seki database is integrated into the search algorithm during game solving.

\begin{comment}
% (overview of this chapter)
The process for creating the seki knowledge table involves two main steps. 
% The process of creating the seki knowledge table involves two main steps. 
Firstly, we will introduce the approach for generating every potential seki pattern within a specified area size. 
% First, we introduce the approach for generating all potential seki patterns within a specified area size. 
After that, all the patterns will be sent to the seki recognizing search. 
% Next, these patterns are processed using the seki recognition search. 
The second part will provide a detailed description of the search algorithm. 
% The second part of the chapter provides a detailed description of the search algorithm. 
In the final section of this chapter, we'll illustrate how to utilize the seki knowledge-table during solving.
% Finally, we demonstrate how to utilize the seki knowledge table during the solving process.
\end{comment}

\subsection{Pattern Enumeration}

\begin{figure}[h]
    \centering
    \begin{tikzpicture}
        % First row (first two images)
        \node at (0, 1.2) {\includegraphics[width=0.18\columnwidth]{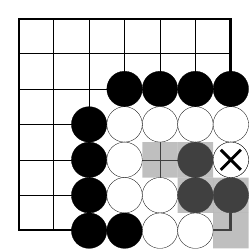}}; % First image
        \node at (2.7, 1.2) {\includegraphics[width=0.18\columnwidth]{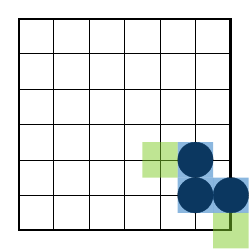}}; % Second image
        \draw[->, thick, line width=0.5mm, black] (1.2, 1.2) -- (1.6, 1.2); % Arrow between first and second images

        % Second row (third and fourth images)
        \node at (0, -1.2) {\includegraphics[width=0.18\columnwidth]{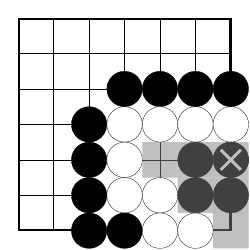}}; % Third image
        \node at (2.7, -1.2) {\includegraphics[width=0.18\columnwidth]{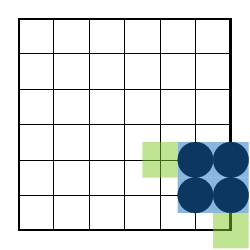}}; % Fourth image
        \draw[->, thick, line width=0.5mm, red] (1.2, -1.2) -- (1.6, -1.2); % Arrow between first and second images

        % Arrows pointing to the central image
        \draw[->, thick, line width=0.5mm, black] (3.9, 1.2) -- (4.3, 0.5);  % From first row to center
        \draw[->, thick, line width=0.5mm, red] (3.9, -1.2) -- (4.3, -0.5); % From second row to center
        \node at (4.2, 1.5)  {\textbf{\tiny Hit}};
        \node at (4.2, -1.5)  {\textcolor{red}{\textbf{\tiny Miss}}};

        % The position of the cylinder is at (7, 0)
        \begin{scope}
            % Draw the top ellipse
            \draw[thick] (5.5, 0.7) ellipse (1 and 0.3);
            % Draw the sides of the cylinder
            \draw[thick] (4.5, 0.7) -- (4.5, -0.7);  % Left side
            \draw[thick] (6.5, 0.7) -- (6.5, -0.7);  % Right side
            % Draw the bottom arc
            \draw[thick] (4.5, -0.7) arc[start angle=180, end angle=360, x radius=1, y radius=0.3];
            % Text in the cylinder
            \node at (5.5, -0.1) {\textbf{\tiny seki database}};
        \end{scope}

        % Bracket surrounding the boxes
        \draw [decorate,decoration={brace,amplitude=10pt,mirror},thick,line width=0.5mm,black](7, 2.3) -- (7, -2.3) node[midway,xshift=1cm]{};

        % database content
        % first row
        \node at (7.7, 1.8) {\includegraphics[width=0.1\columnwidth]{figure/Query_seki_table/sekipattern1.pdf}};
        \node at (9, 1.8) {\includegraphics[width=0.1\columnwidth]{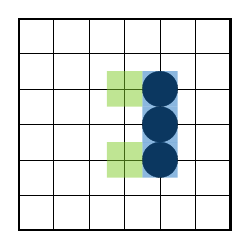}};
        \node at (10.3, 1.8) {\includegraphics[width=0.1\columnwidth]{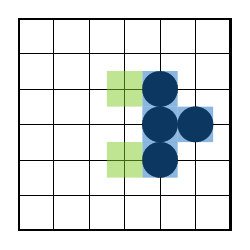}};
        % second row
        \node at (7.7, 0) {\includegraphics[width=0.1\columnwidth]{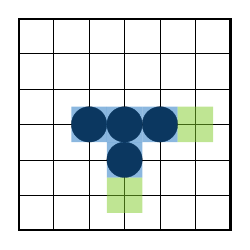}};
        \node at (9, 0) {\textbf{...}};
        \node at (10.3, 0) {\includegraphics[width=0.1\columnwidth]{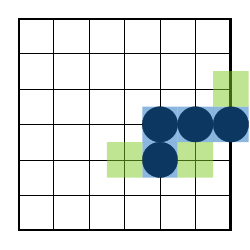}};
        % third row
        \node at (7.7, -1.8) {\includegraphics[width=0.1\columnwidth]{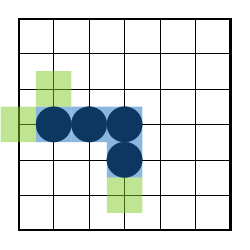}};
        \node at (9, -1.8) {\includegraphics[width=0.1\columnwidth]{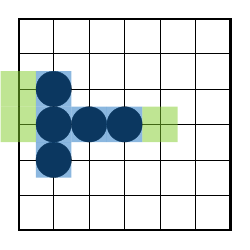}};
        \node at (10.3, -1.8) {\includegraphics[width=0.1\columnwidth]{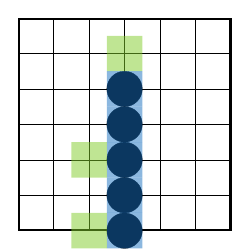}};
        
    \end{tikzpicture}
    \caption{Querying the seki database. The enclosed area is shaded in gray based on the last played move (marked with a cross).}
    \label{fig:seki-table-usage}
\end{figure}

The process of generating a seki database is similar to that of chess endgame tablebases.
As mentioned in subsection \ref{subsec:seki}, we only focus on local seki. 
For all potential local seki patterns, there are three key components: a black boundary, a white block enclosing a contiguous potential seki area, and interior black stones within the seki area. 
There are two examples of such patterns on the left hand side of Fig. \ref{fig:seki-table-usage}.
All potential local seki can be categorized according to the pattern size, just like how chess tablebases are categorized by piece count. In this paper, we enumerate all possible patterns from size 5 to 8.
We skip sizes 4 and below since they are too small to form seki patterns.

We begin by generating all possible contiguous shapes of the specified area size $n$.
For each shape, we create a potential pattern in four steps.
First, we define the generated shape as the seki area.
Next, we surround the area with an enclosing white block.
A black boundary is then added to the pattern to deprive the enclosing white block of all external liberties.
Lastly, we systematically fill the interior area with black stones until there are only two or three empty grids; this will yield ${n \choose 2} + {n \choose 3}$ combinations.

\subsection{Seki Verification and Storage}
For each generated pattern, we mostly follow Niu et al.'s local seki detection method \cite{niu_recognizing_2006} to determine whether they are seki.
As with Niu et al.'s method, we search each pattern twice, where Black and White each play first.
All candidate moves need to be within the seki area.
Moves played outside of the area have no impact, and therefore can be viewed as equivalent to passing; thus, two consecutive passes no longer ends the game.
Following the definition of a seki (see subsection \ref{subsec:seki}), whoever plays inside the area first loses.
For this reason, we prohibit passing as the first move of the search, i.e. the position must change as a result of the first player's first move. 
To avoid perpetual delays, if the position remains the same due to continual passing from both sides, the first player must play to change the situation.
If the pattern inside the area is a seki, the first player is guaranteed to lose.
Following Niu et al.'s method, if both Black and White loses as the first player, the area is a local seki.
Since the seki database is generated offline, with no time constraints, we simply implemented this verification and-or search with depth-first search, instead of the more efficient but elaborate DFPN algorithm.

Next, patterns that are verified to be seki are stored into the database. Since we focus on local seki, we assume that the enclosing white block has no external liberties and eyes. This means we only have to store the contents of each grid (empty or occupied by Black).

\begin{comment}
In this section, we send all the possible seki patterns to the seki recognizing search. 
In reference to Niu et al \cite{niu_recognizing_2006} . 's work, we implement the seki recognizing search for local seki. 
A pattern will be searched twice. The player that goes first is called the Attacker and, otherwise, the defender. 
Black and White will be the attacker alternately. In each round, the pass is prohibited in the attacker’s first move.
That is, the board position must be changed after the first move. 
Given the definition of Seki, once the seki position is changed, the attacker has no chance to win. 
As a result, in each round, if attacker is proven to win, the pattern is not a seki. 
On the other way, attacker is proven to lose in both rounds, the pattern is then proven to be a local seki.
In some cases, the attacker and defender will pass continually. 
To address this issue, we will prohibit the attacker’s next pass move to force the position change. 
The proving process in each round is a proving process of an and-or tree. 
In this work, the search algorithm is implemented by DFS, which can be replaced by other more efficient algorithms such as DFPN.
\end{comment}

\subsection{Using the Seki Database in Solving Killall-Go}
\label{subsec:method_using}
In Killall-Go, the winning condition for White simply requires them to hold any amount of territory. 
This can usually be achieved through UCA, but seki, while rare, can also guarantee life.
Therefore, upon confirmation of either UCA for White or seki, we have reached a terminal position and White's win can be updated accordingly in the and-or tree.

We describe how the seki database is used via Fig. \ref{fig:seki-table-usage}.
To reduce overhead, we only query the seki database if the most recent move is either part of an enclosing White block (as is the case on the top) or within a White enclosed area (as is on the bottom). 
The input for the query consists of the shape of the area (represented by their indices, and denoted by the shaded colors) and whether each grid is empty (green) or contains a black stone (blue).

In this illustrated example, we could not find a matching pattern for the bottom case, which means the search must proceed to obtain the correct game outcome.
On the other hand, the top case is a hit, which means the enclosing white block must be alive due to seki or UCA.
To explain the latter case, keep in mind that we only look for matching patterns inside the enclosed area, which confirms that Black cannot invade successfully. Meanwhile, if White's enclosing block also forms at least one eye, it satisfies the stronger UCA condition. 
In either case, White has secured territory and won.

It is worth noting that there are edge cases of seki that our generation method does not cover.
For example, even in patterns that do not match, an external eye formed by the enclosing white block may form a seki.
Nonetheless, for the game of Killall-Go, we can guarantee that the edge cases cause negligible impact to our search performance.

\section{Experiments}
\begin{comment}
- Experiment
    - Knowledge-Table Information
        -information with different size
    - information of benchmark
        - Time
        - Node count
        - job information
    - dive into unsolved benchmark
        - worker set experiment
        - seki example in the job
\end{comment}

% (setting and outline of experiment)
We perform our experiments on the online fine-tuning solver presented in our previous paper \cite{wu_game_2024}, for which the code is based on the MiniZero framework \cite{wu_minizero_2024}, only changing the top-$k$ configuration from 4 to 2.
Subsection \ref{subsec:table_info} provides statistics related to the generation of the seki database.
The online fine-tuning solver is a distributed solver system that has workers analyzing different positions in parallel.
Subsections \ref{subsec:benchmark} and \ref{subsec:worker_experients} both investigate how the seki database affects performance, where the former looks at the whole solving system, from manager to workers, and the latter looks at job statistics (i.e. only workers).
%Subsection \ref{subsec:benchmark} presents ten Killall-Go openings and compares solving times and node counts when solving these openings with the online fine-tuning solver.
%Subsection \ref{subsec:worker_experients} categorizes jobs according to the percentage of seki encountered during the search, then investigates the solving rate for each category.

\begin{comment}
In the experiment, we follow the online fine-tuning solver in Wu et al. \cite{wu_game_2024}.
Only change the top-k configuration to 2. To show the improvement given by the seki knowledge table, in section 4.2, we prepare a set of benchmarks and utilize the table for solving.
For an unsolved benchmark case without using seki knowledge table, in section 4.3, we collect the job sent by the manager and compare the solving result using worker to show the reason for unsolving the problem.
\end{comment}

\subsection{Database Generation}
\label{subsec:table_info}
\begin{table}[ht]
% \color{blue} % TODO
\caption{Seki database information.}
    \centering
    \setlength{\tabcolsep}{5pt}
    \begin{small}
    \begin{tabular}{crrrr}
    \toprule
    Area size
     & \# Patterns & \# Seki patterns & Seki rate & Time (s) \\
    \midrule
    5 & 28,432    & 1,318  & 4.64\% & 3\\ 
    6 & 133,812   & 8,208  & 6.13\% & 31\\ 
    7 & 578,064   & 51,354 & 8.88\% & 946\\ 
    8 & 2,315,014 & 193,462& 8.36\% &26,716
     \\
    \bottomrule
    5-8 & 3,055,322 & 254,342 & 8.32\% & 27,696
    \\
    \bottomrule
    \end{tabular}
    \end{small}
    \label{tab:table-information}
\end{table}

%(seki table information and comparation of different area size)
% jason tabel fig 格式

We generate seki patterns between area sizes of 5 to 8 using two E5-2683 v3 CPUs, for a total of 16 threads.
Table \ref{tab:table-information} shows the relevant data for each area size, along with the cumulative statistics.
The possible number of patterns roughly increases by four times for each area size increase. 
Larger area sizes mean larger search spaces, and longer times to generate seki entries.
For each increase in area size, the time to generate entries roughly increases 30 fold.
% 5d: change size 7 and 8
The right hand side of Fig. \ref{fig:seki-table-usage} shows eight examples of the patterns stored in the database, two for each area size.
Note that size 8 patterns take up the majority of stored entries in the database.

\subsection{Solver Performance on Benchmark Openings}
\label{subsec:benchmark}

\begin{figure}[h]
    \captionsetup[subfigure]{justification=centering}
    \centering
    \subfloat[A]{
        \includegraphics[width=0.15\columnwidth]{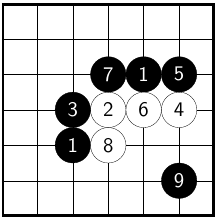}
        \label{fig:seki_opeings_A}
    }
    \subfloat[B]{
        \includegraphics[width=0.15\columnwidth]{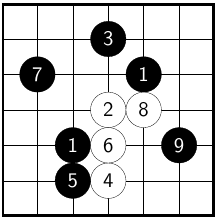}
        \label{fig:seki_opeings_B}
    }
    \subfloat[C]{
        \includegraphics[width=0.15\columnwidth]{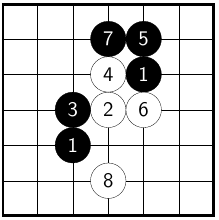}
        \label{fig:seki_opeings_C}
    }
    \subfloat[D]{
        \includegraphics[width=0.15\columnwidth]{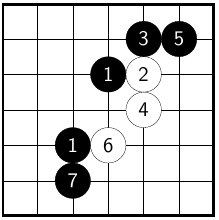}
        \label{fig:seki_opeings_D}
    }
    \subfloat[E]{
        \includegraphics[width=0.15\columnwidth]{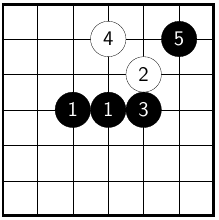}
        \label{fig:seki_opeings_E}
    }
    \\
    \subfloat[F]{
        \includegraphics[width=0.15\columnwidth]{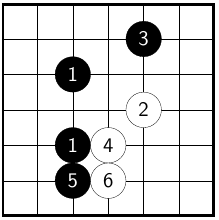}
        \label{fig:seki_opeings_F}
    }
    \subfloat[G]{
        \includegraphics[width=0.15\columnwidth]{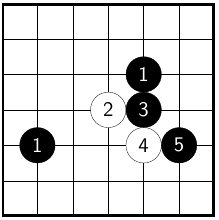}
        \label{fig:seki_opeings_G}
    }
    \subfloat[H]{
        \includegraphics[width=0.15\columnwidth]{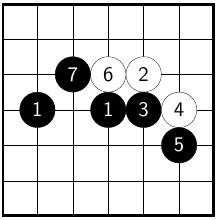}
        \label{fig:seki_opeings_H}
    }
    \subfloat[Opening1]{
        \includegraphics[width=0.15\columnwidth]{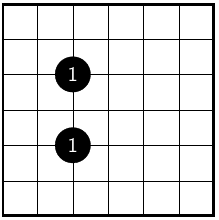}
        \label{fig:seki_opeings_1}
    }
    \subfloat[Opening2]{
        \includegraphics[width=0.15\columnwidth]{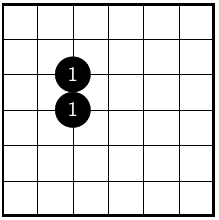}
        \label{fig:seki_opeings_2}
    }
    \caption{The collection of openings used to evaluate the seki database.}
    \label{fig:openings}
\end{figure}

% \begin{figure}[!t]
% \centering
% \includegraphics[width=0.9\columnwidth]{figure/test_benchmark.png}
% \caption{Benchmark of Experiment}
% \end{figure}
% \subsection{Experiment w/ w/o Seki table in manager}
\begin{table}
% \color{blue} % TODO
\caption{Solving results for 10 7x7 Killall-Go benchmark openings.}
    \centering
    \setlength{\tabcolsep}{3pt}
    \begin{adjustbox}{width=\columnwidth}
    \begin{tabular}{crrrrrrrr|rr}
    \toprule
     &\multicolumn{3}{c}{w/o Seki table}& & \multicolumn{3}{c}{w/ Seki table}&& \multicolumn{2}{c}{Reduction rate(\%)} \\
     \cline{2-4}
     \cline{6-8}
     
     & Time(s) & \# Nodes & \makecell[c]{Avg. \\jobs time(s)} & 
     & Time(s) & \# Nodes &  \makecell[c]{Avg. \\jobs time(s)} &&\makecell[c]{Time}&\makecell[c]{Nodes} \\
    \midrule
    A& $\ge$86,400 & - & 246.86& &  \textbf{482} & 11,055,902& 4.89&& -& -\\
    
    B& $\ge$86,400 & -& 59.83& & \textbf{5,719}& 284,020,298& 24.67&& -& -\\ 
    \midrule
    C& 14,100& 467,441,600& 67.74& & \textbf{11,257}& 581,523,050& 38.27&& 20.16\% & -24.41\% \\
    D& 68,582& 3,021,039,537& 41.39& &\textbf{30,660}& 1,441,063,897& 25.70&& 55.29\%& 52.30\%\\
    E& 710& 24,483,162& 7.20& & \textbf{624}& 22,256,876& 7.07&& 12.06\%& 9.09\%\\
    F& $\ge$86,400 & -& 40.65& &\textbf{33,761}& 1,479,511,498& 21.04&& -& -\\
    G& 21,744& 1,057,881,380& 30.23& &\textbf{13,288}& 706,327,784& 27.05&& 38.89\%& 33.23\%\\
    H& 1,223& 56,350,104& 26.90& &\textbf{927}& 49,982,460& 20.87&& 24.23\%& 11.30\%\\
    \midrule
    Opening1& 14,641& 767,298,193& 25.96& & \textbf{13,122}& 682,063,159& 24.84&& 10.38\%& 11.10\%\\
    Opening2& 30,240& 1,654,756,361& 25.31& & \textbf{24,683}& 1,376,817,358& 21.72&& 18.38\%& 16.80\%\\
    \bottomrule
    \end{tabular}
    \label{tab:manager-experiment}
    \end{adjustbox}
\end{table}

\begin{comment}
(result of solving result
    -benchmark is depart as theree part
        -A~B go professional player suggest
        -C~H from self-play
        -opening12 baseic opening
)
\end{comment}

We now try to solve a collection of ten openings using our previously presented online fine-tuning solver \cite{wu_game_2024}.
The benchmark problems can be separated into three parts.
Cases A and B are problems suggested by Go experts, with a high probability of seki occurring.
Cases C to H are problems that were collected during self-play training for our deep learning-based heuristic.
Lastly, openings 1 and 2 are frequently used opening moves in Killall-Go. In other words, they are typical use cases when trying to solve Killall-Go.

% (A-B)

Table \ref{tab:manager-experiment} shows the results of solving each case.
First, when not using the seki database, cases A and B cannot be solved within a day. 
With the seki database, they can be solved in 482 and 5,719 seconds, respectively. 
This demonstrates that when seki is inevitable, it is significantly more costly, even infeasible, to analyze without some kind of seki detection method. 
% For these problems, the jobs sent from the manager also contain seki situations.
In a distributed game solver (see subsection \ref{subsec:background_solver}), the manager sends interesting positions (jobs) to workers to analyze in parallel.
In Table \ref{tab:manager-experiment}, the average job time indicates how much time each worker spends analyzing these interesting positions.
For case A, the average job time is 246.86 seconds without seki, but 4.89 seconds after using the database.
Note that unsolved jobs will take roughly 420 seconds.
This implies that workers might be stuck in long sequences of capturing and re-capturing. 
We perform additional experiments to analyze jobs in subsection \ref{subsec:worker_experients}.

% (C-H, table is also useful in benson case)
% ting: check to see the benchmark order in the table matches fig 3
With the exception of case F, problems C to H are solvable even without the seki database. However, using it yields a 20\% to 50\% discount on solving time, solving nodes, and jobs average time. 
Only in case C were there a 24.4\% increase in total nodes searched.  
This was caused by the manager sending more jobs due to a significant discount on the average job time.
In addition, we examined the search logs and discovered that a matching pattern could indicate either a seki or the stronger UCA requirement, as explained in subsection \ref{subsec:method_using}. 
In Killall-Go, both seki and UCA indicates a White win.  
This allows us to skip the Benson algorithm completely, significantly reducing the time and nodes necessary to solve the position.
% (opening12)
Similarly, utilizing the seki database for openings 1 and 2 also gives a discount of 10.38\% and 18.38\% for time, respectively.
This shows that the seki database can still be useful when solving typical openings, where seki may or may not be part of the solution.

\subsection{The Seki Database's Impact on Job Solve Rates}
\label{subsec:worker_experients}
% (experiment setting, background of jobs)

In subsection \ref{subsec:benchmark}, we looked at the seki database's impact on the online fine-tuning solver holistically.
In this subsection, we now turn to its impact on individual jobs, categorized by the number of seki encountered during the job.
We randomly sampled 10,000 jobs sent from the manager, while not using the seki database when solving opening A, as shown in Fig. \ref{fig:seki_opeings_A}.
These jobs are then recalculated with and without the seki database, then categorized by the seki database hit rate within each job, where the hit rate is calculated by the number of matching patterns divided by the total number of terminal nodes encountered while analyzing the job.
As an extreme example, if the position shown in the upper left corner of Fig. \ref{fig:seki-table-usage} is sent as a job to a worker, it will match an entry in the database, and recognized as a win for White, with a seki hit rate of 100\%.
Alternatively, if the position in the bottom left is sent as a job, no matches can be found, and the search will proceed as usual. If it is then solved and exactly two positions are matched among 100 terminal nodes, its hit rate is 2\%.

% (result of the experiment)
\begin{table}
% \color{blue} % TODO
\caption{Solving rate of opening A.}
    \centering
    \setlength{\tabcolsep}{5pt}\
    \begin{small}
    \begin{tabular}{crrr}
    \toprule
    \makecell[c]{Seki table \\ hit rate } & \# jobs &\multicolumn{1}{c}{w/ Seki table}& \multicolumn{1}{c}{w/o Seki table} \\

    \midrule
    0\%& 4,580& \multicolumn{2}{c}{94.13\%}\\
    $(0\%-10\%)$& 3,856& 85.68\% & 33.87\%\\
    $[10\%-50\%)$& 389 & 91.77\% & 6.68\%\\
    $[50\%-100\%]$& 1,175 & 100.00\%& 3.40\%\\
    \bottomrule
    \end{tabular}
    \end{small}
    \label{tab:caseAjob-worker}
\end{table}

In Table \ref{tab:caseAjob-worker}, jobs with 0 hit rate have a 94.13\% solving rate, and the seki database does not improve the solving rate.
However, with only 10\% hit rate, the solving rate without using the seki database drops drastically to 33.87\%. 
Where the hit rate exceeds 10\%, the solving rate without using the seki database drops to less than 6.68\%.
In contrast, when using the database, the solving rate is higher than 85\% in all cases. 
This shows that when seki are possible, the database can be tremendously helpful.

\section{Conclusion}
\begin{comment}
- conclusion: use seki table is helpful ...
    - Seki table can help reduce the solving complexity of seki situation, can get a time discount more than 80%
    - When it comes to normal situation (not a seki). Seki table can also boost the efficiency by mathching at least seki situaiton, and thus reduce the solving time by 10~20%.
- limitation: our method only find some specific type of seki
- future work: build a bigger knowledge table by including more knowledge, such as seki with open boundary, global seki table, rzone, one-eye table, ....
This concept can be applied to other applications.
\end{comment}

This paper clearly illustrates that attempting to solve 7x7 Killall-Go without seki detection is prohibitively costly even for simple positions that may encounter relatively few seki situations. 
When encountering positions where seki appears more than 10\% of the time, the solving rate drops to lower than 6.68\%. 
%With a seki database, our solver treats seki as terminal positions, maintaining a solving rate above 91.77\%. 
In the most extreme case, subsection \ref{subsec:benchmark} demonstrates that previously unsolvable seki positions can now be solved in just 482 seconds, especially since it avoids exhaustive seki detection algorithms during runtime. 

Even for common openings in Killall-Go, seki knowledge also gives a 10-20\% discount on solving time and nodes. 
Other than local seki, we could also extend the database for global seki, edge cases, or relevancy zones \cite{shih_novel_2022}. 
We believe that the underlying concept of endgame databases such as the database presented in this paper can also be applied to other applications.

\begin{credits}
\subsubsection{\ackname}

This research is partially supported by the National Science and Technology Council (NSTC) of the Republic of China (Taiwan) under Grant Numbers 111-2222-E-001-001-MY2 and 113-2221-E-001-009-MY3.

\end{credits}

\bibliographystyle{splncs04}
\bibliography{references}

\end{document}